\documentclass[english]{lni}
\newtheorem{defn}{Definition}[section]  % Definitionen
\usepackage{amsmath}
\usepackage{bbm}
\usepackage{booktabs}    % für \toprule, \midrule, \bottomrule
\usepackage{array}       % für benutzerdefinierte Spaltenbreiten (z.B. p{3.2cm})
\usepackage{caption}     % für \caption* bei Tabellen ohne Nummerierung

\usepackage{booktabs}      % fürTabellen
\usepackage{pdflscape}     % für Querformatseiten
\usepackage[toc,page]{appendix} % für Anhang mit TOC-Eintrag
\usepackage{caption}       % für flexible Beschriftung
\usepackage{longtable}     % falls Tabellen über mehrere Seiten gehen
\usepackage{geometry}
\usepackage{adjustbox}

\begin{document}
%%% Mehrere Autoren werden durch \and voneinander getrennt.
%%% Die Fußnote enthält die Adresse sowie eine E-Mail-Adresse.
%%% Das optionale Argument (sofern angegeben) wird für die Kopfzeile verwendet.
\title[Variance-Aware Bandit Algorithms]{A Framework for Fair Evaluation of Variance-Aware Bandit Algorithms}
 \author[1]{Elise Wolf}{elise.marie.wolf@students.uni-mannheim.de}{0009-0000-5130-4352}
\affil[1]{University of Mannheim\\School of Business Informatics and Mathematics\\B 6, 26\\68159 Mannheim\\Germany}
\maketitle

\begin{abstract}
Multi-armed bandit (MAB) problems serve as a fundamental building block for more complex reinforcement learning algorithms. However, evaluating and comparing MAB algorithms remains challenging due to the lack of standardized conditions and replicability. This is particularly problematic for variance-aware extensions of classical methods like UCB, whose performance can heavily depend on the underlying environment. In this study, we address how performance differences between bandit algorithms can be reliably observed, and under what conditions variance-aware algorithms outperform classical ones. We present a reproducible evaluation designed to systematically compare eight classical and variance-aware MAB algorithms. The evaluation framework, implemented in our \textit{Bandit Playground} codebase, features clearly defined experimental setups, multiple performance metrics (reward, regret, reward distribution, value-at-risk, and action optimality), and an interactive evaluation interface that supports consistent and transparent analysis. We show that variance-aware algorithms can offer advantages in settings with high uncertainty where the difficulty arises from subtle differences between arm rewards. In contrast, classical algorithms often perform equally well or better in more separable scenarios or if fine-tuned extensively. Our contributions are twofold: (1) a framework for systematic evaluation of MAB algorithms, and (2) insights into the conditions under which variance-aware approaches outperform their classical counterparts.
\end{abstract}


\section{Introduction}

MAB problems provide a basis for decision-making under uncertainty in reinforcement learning, where agents must balance exploration of unknown actions with exploitation of known high-reward options \cite{sutton}.

Among the numerous algorithmic approaches to this problem, classical methods such as the Upper Confidence Bound (UCB) algorithm \cite{auer2002finite} offer robust strategies based on confidence bounds. More extensions like UCB-V \cite{audibert2} incorporate variance estimates into these bounds to better handle environments where expected rewards are similar but variances differ. While variance-aware algorithms are theoretically promising, a reliable and comprehensive analysis of their advantages over classical methods remain underexplored.

Despite the growing number of proposed algorithms, the field lacks a comprehensive framework for reproducible, empirical comparison. Existing studies often vary in setup, focus on a narrow set of metrics, specify application domains, or omit reproducibility. For example, Audibert et al. \cite{audibert2} demonstrated UCB-V’s advantages in high-stochasticity environments but compared only a limited number of algorithms under narrowly defined conditions. Similarly, Auer et al. \cite{auer2002finite} introduced UCB-Tuned and compared it to $\varepsilon$-Greedy in basic Bernoulli environments, yet did not provide reproducible setups. The most extensive comparison to date \cite{Mukherjee} evaluates a larger set of algorithms, including EUCBV, but uses fixed scenarios without variation across environment parameters or replicability. To our knowledge, no existing work offers a systematic, reproducible evaluation that contrasts variance-aware and non-variance-aware algorithms across diverse conditions. Our bottom-up analysis approach aims to abstract from domain-specific assumptions and focus on controlled scenarios, addressing the existing gap.

The overarching research questions guiding this work are as follows:

\begin{itemize}
    \item[1.] How can algorithm performance differences be reliably detected and quantified using a reproducible evaluation within a standardized framework?
    \item[2.] Under which conditions do variance-aware MAB algorithms outperform classical and non-variance-aware counterparts? 
\end{itemize}

Our contribution is twofold: First, we provide a structured, extensible implementation of eight widely discussed algorithms. Those are split into 
\begin{itemize}
    \item standard algorithms, including Explore-Then-Commit (ETC), $\varepsilon$-Greedy, and UCB algorithm,
    \item variance-aware algorithms, including UCB-V, UCB-Tuned and Efficient-UCBV (EUCBV),
    \item and other non-variance-aware algorithms, including Probability Approximately Correct UCB (PAC-UCB) and UCB-Improved.
\end{itemize}

Second, we perform a systematic empirical comparison across representative scenarios, highlighting the conditions under which variance-aware methods exhibit superior performance.

This study aims to support reproducibility and fair algorithmic comparison. To this end, we publish our results along with detailed instructions for replicable use via our open science codebase, the Bandit Playground. The codebase and associated resources are publically available at: \url{https://github.com/eelisee/bandit_playground/tree/v1.0-SKILL2025}.

The remainder of this paper is structured as follows: Section~\ref{sec:background} reviews the theoretical foundations of the implemented algorithms. To address Research Question 1, we developed a simulation framework in Section~\ref{sec:methodology} together with the setup for the three evaluation scenarios used to test algorithm performance, algorithm parameterization, evaluation metrics, and the visualization-based evaluation framework. In Section~\ref{sec:results}, we address Research Question~2 through a comparative analysis across these scenarios. Finally, Section~\ref{sec:discussion} discusses key findings, highlights limitations, and reflects on threats to validity.

\section{Background and Related Work}
\label{sec:background}

\subsection{The Stochastic Multi-Armed Bandit Problem}

The stochastic MAB problem models a sequential decision-making scenario under uncertainty and was first introduced by Thompson in 1933~\cite{thompson1933}. At each time step $t \in \{1, \dots, n\}$, a learner selects an action (or \emph{arm}) $A_t \in \mathcal{A} = \{a_1, \ldots, a_K\}$ and receives a stochastic reward $X_t \sim \mathbb{P}_{A_t}$. The goal is to maximize the total reward over $n$ rounds (the \emph{horizon}) or, equivalently, to minimize the cumulative \emph{regret} compared to always pulling the best arm \cite{lattimore}.

Each arm $a_k$ has an unknown expected reward $Q_{a_k} = \mathbb{E}[X_{a_k}]$, and the optimal arm is defined as $a^* = \arg\max_{a_k} Q_{a_k}$ with value $Q^* = Q_{a^*}$. The suboptimality gap of arm $a_k$ is denoted by $\Delta_{a_k} = Q^* - Q_{a_k}$, and $T_{a_k}(t) = \sum_{s=1}^{t} \mathbbm{1}_{A_s = a_k}$ counts how often arm $a_k$ has been played up to time $t$ \cite{lattimore}.

\begin{defn}[Regret, {\cite{lattimore}}]
The cumulative regret up to round $n$ is defined as $R_n = n Q^* - \mathbb{E}\left[\sum_{t=1}^n X_t\right],$
measuring the loss incurred by playing suboptimal arms.
\end{defn}

\begin{defn}[Estimated Action Value, {\cite{sutton}}]
The estimated action value of arm $a_k$ at time $t$ is $\hat{Q}_{a_k}(t) = \frac{1}{T_{a_k}(t)} \sum_{s=1}^{t} X_s \cdot \mathbbm{1}_{A_s = a_k},$
i.e. the empirical average reward from selecting arm $a_k$.
\end{defn}

\subsection{Classical Algorithms}

\paragraph{Explore-Then-Commit (ETC).}
This algorithm explores all arms uniformly for a fixed number of steps $m$ and then commits to the empirically best arm $\hat{Q}_{a_k}$ for the rest of the horizon. While simple, its performance heavily depends on a good choice of $m$, which requires knowledge of $\Delta_{a_k}$ or the horizon $n$~\cite{lattimore}.

\paragraph{$\varepsilon$-Greedy.}
With probability $\varepsilon$, a random arm is selected (exploration); otherwise, the empirically best arm is chosen (exploitation). Although it guarantees a minimum level of exploration, the constant exploration rate leads to suboptimal long-term performance~\cite{sutton}.

\paragraph{UCB.}
UCB~\cite{auer2002finite} balances exploration and exploitation by selecting the arm with the highest upper confidence bound
\[
UCB_{a_k}(t) = \hat{Q}_{a_k}(t) + \sqrt{\frac{2 \log t}{T_{a_k}(t)}}.
\]
The idea is that the true mean lies below this bound with high probability. UCB favours actions with high uncertainty and achieves logarithmic regret~\cite{auer2002finite}.

\subsection{Variance-Aware Algorithms}

\paragraph{UCB-V.}
UCB-V~\cite{audibert2} extends UCB by incorporating an empirical estimate of the reward variance $\hat{\sigma}^2_{a_k}(t)$ into the confidence bound:
\[
B_{a_k, T_{a_k}}(t) = \hat{Q}_{a_k}(t) + \sqrt{\frac{2 \hat{\sigma}^2_{a_k}(t) \cdot \varepsilon_{T_{a_k},t}}{T_{a_k}(t)}} + c \cdot \frac{3b \varepsilon_{T_{a_k},t}}{T_{a_k}(t)}.
\]
The exploration function $\varepsilon_{T_{a_k},t} = \theta \log(t)$ is a typical choice and increases over time for an arm $a_k$, if it has not been selected for a while, eventually dominating $\hat{Q}_{a_k}$~\cite{audibert2}. The bound depends on parameters $\theta$, $b$, and $c$, which control the strength and shape of the exploration term.

\paragraph{UCB-Tuned.}
UCB-Tuned~\cite{auer2002finite} further refines UCB by adjusting the confidence bound based on the empirical variance. While no formal regret bound was shown, the algorithm performed better than UCB in the experiments conducted in the original study.

\paragraph{EUCBV.}
Introduced by Mukherjee et al.~\cite{Mukherjee}, EUCBV combines the variance-adaptive confidence intervals from UCB-V with an arm elimination strategy. It uses parameters $\rho$ and $\psi$ to control the aggressiveness of elimination and the scaling of exploration over time. We set $\psi=n / K^2$.

\subsection{Non-Variance-Aware UCB-Based Algorithms}

\paragraph{PAC-UCB.}
PAC-UCB, introduced in Audibert et al.~\cite{audibert2} modifies the UCB-V algorithm where the exploration function is solely dependent on $T_{a_k}(t)$, typically chosen as $\varepsilon_{T_{a_k}(t)}= \log{\{K s^q \beta^{-1}\}} \vee 2$. This results in variance-independent confidence intervals.

\paragraph{UCB-Improved.}
UCB-Improved~\cite{auerortner} uses adaptive confidence bounds and a phased elimination process. After each phase, arms that are statistically unlikely to be optimal are removed. The algorithm halves the estimate of $\delta$ iteratively to progressively refine the best arm.

\section{Methodology}
\label{sec:methodology}

\subsection{Design Objectives and Constraints}
\label{sec:design-objectives}

This study aims to provide a controlled, interpretable, and reproducible comparison of classical and variance-aware MAB algorithms. The experimental design is guided by the central objective of prioritizing comparability across algorithms by enforcing unified simulation semantics. All algorithms are evaluated under identical initialization, deterministic random seeds, and normalized time progression (one action per time step). This ensures that observed differences in performance stem from algorithmic principles rather than structural or procedural artifacts.

We deliberately restrict our evaluation to two-armed scenarios to allow for clearer interpretability of behavioral patterns, convergence speed, and risk trade-offs, although the framework supports up to three-armed settings. Constraints of the experimental design include limited computational resources and the need to keep data volume manageable. As such, the evaluation avoids large-scale hyperparameter tuning or extensive high-dimensional environments. Instead, it focuses on a curated but diverse set of problem instances and algorithm classes. Similarly, the number of checkpoints for data recording is intentionally limited to reduce memory overhead without sacrificing analytical granularity.

\subsection{Experimental Setup}
\label{sec:experimental-setup}
All experiments use $K$-armed bandit environments with independent Bernoulli rewards, i.e., $X_{a_k, t} \sim \text{Ber}(p_k)$. These binary, i.i.d. rewards reflect a standard assumption in classical and variance-aware algorithms, enabling controlled analysis of mean and variance effects without added complexity. We restrict our study to two-armed settings ($K=2$) to isolate subtle performance differences, aligning our experimental setup with our bottom-up analysis approach. To reflect varying levels of problem difficulty, we define three evaluation scenarios:
\begin{itemize}
    \item \textbf{Scenario A (Baseline)}: $p_1 = 0.8$, $p_2 = 0.9$. This clear-gap setting provides a baseline for learning behavior under easily distinguishable arms.
    
    \item \textbf{Scenario B (Low-Variance Micro-Gap)}: $p_1 = 0.895$, $p_2 = 0.9$, yielding a small mean gap with low variance ($Var(X_1) \approx 0.093975$, $Var(X_2) = 0.09$), to evaluate the ability to distinguish between nearly equivalent arms with high statistical similarity.
    
    \item \textbf{Scenario C (High-Variance Micro-Gap)}: $p_1 = 0.89$, $p_2 = 0.895$. Despite a similarly small gap, the slightly higher uncertainty ($Var(X_1) \approx 0.0979$, $Var(X_2) \approx 0.093975$) introduces more noise, challenging the algorithms’ robustness to uncertainty.
\end{itemize}

The choice of $p_k$ values from the set $\{0.8, 0.89, 0.895, 0.9\}$ is based on the need to simulate realistic yet analytically traceable learning environments. Values close to 0.9 maximize the reward ceiling while allowing subtle performance differentiation. Mean gaps between 0.005 and 0.1 are chosen to distinguish between easy and hard exploration regimes.

Each algorithm is executed for $T = 1{,}000{,}000$ steps, referring to the horizon, in each scenario and repeated across $100$ independent trials to ensure statistical validity.

\subsection{Algorithms and Parameter Settings}
\label{sec:algorithms-and-parameter-settings}

The study includes eight algorithms representing classical, non-variance-aware, and variance-aware strategies. Parameter configurations are selected to cover both standard defaults as given in the original paper of these algorithms and practically relevant variations. Table~\ref{tab:algos} provides an overview of the implemented algorithms along with their tuning spaces. 

The extent of hyperparameter tuning varies intentionally across algorithms. For widely used baselines such as ETC and $\varepsilon$-Greedy, we explore multiple values to examine behavior under over- and under-exploration. In contrast, algorithms with either well-established default settings or no tunable parameters are evaluated as-is to maintain focus on interpretable comparisons without introducing excessive configuration variance.

\begin{table}[h]
\centering
\caption{Evaluated algorithms, categories, hyperparameters, and sources.}
\begin{tabular}{llll}
\toprule
\textbf{Algorithm} & \textbf{Category} & \textbf{Hyperparameter Configuration} & \textbf{Source} \\
\midrule
ETC& Standard & $m=10, 100, 1000, 10^4, 10^5$ & \cite{lattimore} \\
$\varepsilon$-Greedy & Standard & $\varepsilon =0.5, 0.1, 0.05, 0.01, 0.005$ & \cite{lattimore} \\
UCB & Standard & -- & \cite{auer2002finite} \\
UCB-Tuned & Variance-aware & -- & \cite{auer2002finite} \\
UCB-V & Variance-aware & $\{\theta=1,\ c=1,\ b=1\}$ & \cite{audibert2} \\
EUCBV & Variance-aware & $\{\rho=0.5\}$ & \cite{Mukherjee} \\
PAC-UCB & Non-variance-aware & $\{c=1,\ b=1,\ q=1.3,\ 
\beta=0.05\}$ & \cite{audibert2} \\
UCB-Improved & Non-variance-aware & $\{\delta=1\}$ & \cite{auer2002finite} \\
\bottomrule
\end{tabular}
\label{tab:algos}
\end{table}

\subsection{Design Measures for Eliminating Systematic Bias}

To prevent structural artifacts from affecting algorithm performance comparisons, several measures are taken to eliminate implementation-related systematic bias. 

Firstly, all algorithms were initialized using the same scheme for arm counts and empirical means to avoid advantages during learning and evaluation phases that might arise from algorithmic specific priors. 

Secondly, random number generation is fully deterministic via fixed seeds, ensuring experiment reproducibility. 

Thirdly, to eliminate order-based biases (e.g., arising from deterministic tie-breaking in $\texttt{argmax}$), we evaluate all permutations of arm index ordering. This prevents structural favouritism of one arm over another based solely on its position in the action vector. 

Lastly, algorithm-specific behaviors that might introduce unfair advantages are explicitly standardized. For example, the $\varepsilon$-Greedy algorithm is evaluated across a range of $\varepsilon$ values to account for both over- and under-exploration, which could otherwise distort long-term performance comparisons. Specifically, for the UCB-Improved algorithm, we adapted the original elimination procedure, where multiple arms from the set of remaining arms may be selected within a single time step, by enforcing that only one arm is pulled per time step. This modification ensures consistency in time step semantics across all evaluated algorithms, thereby avoiding structural unfairness in comparison.

\subsection{Performance Metrics and Data Recording}
\label{sec:metrics-recording}

For each simulation run, a structured set of primary metrics is recorded at predefined time steps. These metrics provide the empirical foundation for all quantitative and visual analyses. Each algorithm–configuration pair yields two datasets: (1) raw per-run results and (2) averaged metrics across the 100 independent trials.

The recorded primary metrics per run include:
\begin{itemize}
    \item \textbf{Cumulative reward:} $\sum_{k=1}^{t} X_k$
    \item \textbf{Cumulative regret:} $R_t = \sum_{k=1}^{t} (Q^* - Q_{A_k})$
    \item \textbf{Suboptimal arm pulls:} Count of actions $A_k$ with $Q_{A_k} < Q^*$
    \item \textbf{Binary reward statistics:} Total number of rewards with values 0 and 1
\end{itemize}

Checkpoints are chosen logarithmically to capture early dynamics and long-term trends, $t \in \{2, 3, 100, 200, 2\cdot10^3, 10^4, \dots, 10^6\}$. All primary metrics are rounded to two decimal places to avoid artifacts from floating-point precision and to facilitate comparison across algorithms. Beyond classical regret-based metrics, we also compute the Value at Risk $\textit{VaR}_\alpha(R_t)$ for each configuration. This complementary metric captures worst-case performance quantiles by evaluating the distribution of cumulative rewards across runs~\cite{audibert2}.

\subsection{Standardized Visualization-Based Evaluation Framework}

To ensure reliable and reproducible comparison of MAB algorithms, we developed a structured evaluation framework that integrates a set of six complementary visual analysis. Table~\ref{tab:visualizations} summarizes each visualization’s analytical focus. All simulation outputs are logged in structured CSV formats to allow exact reconstruction of experiments and plots. This design directly supports the reproducibility objectives defined in Research Question~1.

The evaluation procedure is parameterized to allow for a systematic variation of experimental settings. Specifically, three aspects of the evaluation can be dynamically configured prior to each simulation batch: (1) the mean reward probabilities $p_k$ of the arms $a_1$, $a_2$, and if a three-armed setting is defined, optionally $a_3$, which directly influence all performance metrics; (2) the selection of the algorithm under evaluation, particularly relevant for assessing final regret distributional properties; and (3) the confidence level $\alpha \in \{0.01, 0.05, 0.1\}$ used in the Value-at-Risk analysis.

\begin{table}[h]
\centering
\begin{tabular}{>{\raggedright\arraybackslash}p{3cm} >{\raggedright\arraybackslash}p{3.5cm} >{\raggedright\arraybackslash}p{4.6cm}}
\toprule
\textbf{Metric Visualization} & \textbf{Measured Quantity} & \textbf{Analytical Insight} \\
\midrule
Average Total Reward Over Time & Cumulative reward over time $\sum_{k=1}^{t} X_k$ & Overall reward-maximizing performance and convergence behavior. \\
\midrule
Average Regret Over Time & Cumulative regret over time $R_t$ & Speed and stability of convergence toward the optimal policy. \\
\midrule
Reward Outcome Distribution & Count of rewards 0 and 1 & Empirical reward certainty and frequency of successful actions. \\
\midrule
Distribution of Total Regret & Regret values $R_{1{,}000{,}000}$ & Performance stability, spread, and skewness across simulations. \\
\midrule
Value-at-Risk Analysis & Quantiles of regret distribution $\textit{VaR}_\alpha(R_{1{,}000{,}000})$ & Risk-sensitive comparison using $\textit{VaR}_\alpha$ at different confidence levels. \\
\midrule
Suboptimal Action Ratio Over Time & Ratio of suboptimal choices to time & Convergence toward optimal arm selection by monitoring suboptimal choices. \\
\bottomrule
\end{tabular}
\caption{Overview of visualizations used for algorithm evaluation. Each view supports a distinct perspective on performance and risk behavior.}
\label{tab:visualizations}
\end{table}

\section{Results}
\label{sec:results}

This section addresses Research Question 2. To this end, we compare algorithm performance across three representative evaluation scenarios that differ in the distribution of arm means and variances.

\subsection{Scenario A – Baseline ($p_1=0.8$ vs. $p_2=0.9$)}
\label{sec:results-scenario-a}

Scenario A represents a two-armed Bernoulli bandit problem with a reward gap of $\Delta_2^A = 0.1$. This environment enables rapid arm differentiation and thus serves as a baseline to assess exploitation efficiency. Table~\ref{tab:scenario-a-summary} presents aggregated results across 100 runs. To assess the stability of each algorithm, we report $p$-values from a $\chi^2$-test comparing each algorithm's reward variance to the expected variance at T=1{,}000{,}000 of the arm with highest mean reward.

ETC with 100 exploration rounds performs best, achieving minimal regret ($10.1$) and variance. Its small suboptimal pull rate confirms that a short, well-tuned initial exploration phase suffices in this environment.

Variance-aware algorithms like UCB-Tuned (regret $30.7$) and EUCBV (regret $47.7$) also perform strongly, exhibit negligible variance and minimal suboptimal arm selection ($0.00031$ and $0.00048$). While slightly worse than ETC with 100 exploration rounds, they adapt efficiently without requiring parameter tuning.

PAC-UCB and UCB-V rank slightly lower (regret $\sim100$), converging more slowly despite the large gap. Their performance remains solid but not optimal.

In contrast, over- or underexploration severely hurts performance. ETC with excessive exploration (10{,}000 rounds) or highly exploratory Greedy strategies ($\varepsilon=0.5$) incur high regret ($\geq 1{,}000$). UCB-Improved performs worst overall, with extreme regret (32{,}500.11) and instability (suboptimal rate 0.32500).

It is important to contextualize these findings regarding algorithm tuning. While ETC with 100 exploration rounds achieves the best performance, it relies on knowledge of an effective exploration length, which is typically unavailable in practice. In contrast, UCB-Tuned and EUCBV do not depend on environment-specific parameters and still perform near-optimally. Therefore, the apparent superiority of tuned algorithms like ETC must be interpreted in light of their practical feasibility.

In summary, this scenario demonstrates that in simple, stationary environments with a large reward gap, algorithms that either use well-calibrated early exploration (e.g., ETC) or incorporate adaptive confidence bounds mainly via variance-awareness (e.g., UCB-Tuned, EUCBV) are particularly effective.

\begin{table}[h]
\centering
\begin{tabular}{lrrrrr}
\toprule
\textbf{Algorithm} & \textbf{Average Regret} & \textbf{Reward Variance} & \textbf{Subopt. Ratio} & \textbf{p-value} \\
\midrule
ETC ($m=100$) & \textbf{10.10} & 78,844.93 & 0.00010 & 1.00 \\
ETC ($m=1,000$) & 100.00 & 78,385.97 & 0.00100 & 1.00 \\
UCB & 238.17 & 78,515.43 & 0.00238 & 1.00 \\
Greedy ($\varepsilon=0.005$) & 422.99 & 106,607.54	 & 0.00423 & 0.00 \\
ETC ($m=10,000$) & 1,000.00 & 77,952.99 & 0.01000 & 1.00\\
Greedy ($\varepsilon=0.5$) & 25,002.20 & 99,490.43 & 0.25002 & 0.00 \\
\hline
UCB-Tuned & \underline{30.67} & 78,995.41 & 0.00031 & 1.00 \\
EUCBV & 47.73 & 79,188.56 & 0.00048 & 1.00 \\
UCB-V & 109.93 & 78,689.64 & 0.00110 & 1.00 \\
\hline
PAC-UCB & 99.50 & 79,232.96 & 0.00099 & 1.00 \\
UCB-Improved & 32,500.11 & 2,143,278,875.08 & 0.32500 & 0.00 \\
\bottomrule
\end{tabular}
\caption{Scenario A - Summary Statistics across 100 Runs ($T=10^6$)}
\label{tab:scenario-a-summary}
\end{table}

\subsection{Scenario B – Low-Variance Micro-Gap ($p_1 = 0.895$ vs.\ $p_2 = 0.9$)}
\label{sec:results-scenario-b}

Scenario B examines a setting with minimal variance and a narrow reward gap of $\Delta_2^B = 0.005$ between arms. This environment challenges algorithms to detect subtle differences with high confidence and tests whether variance-aware methods can capitalize on the statistical stability of the arms.

Table~\ref{tab:scenario-b-summary} presents the results. The best-performing algorithm is again the heavily tuned ETC with 10,000 exploration steps, which achieves the lowest regret (150.66) but high variance (510,772.0). However, its performance relies on ideal calibration of the exploration budget, limiting generalizability.

Among the UCB variants, UCB-Tuned performs particularly well (regret $212.21$), while the standard UCB underperforms significantly (regret $1,127.17$), lacking mechanisms to adapt to variance and thus struggling with fine-grained reward differences, suggesting that variance sensitivity offers advantages in settings with minimal reward gap between arms.

UCB-Improved again performs poorly (regret $2,525.00$), reinforcing the pattern that its conservative bounds and prolonged exploration phases are poorly suited for stable environments.

$\chi^2$ tests confirm that most algorithms exhibit significant differences in reward variance. UCB-Tuned and Greedy with $\varepsilon=0.5$ match the stability level (p-value = 1.0), while others (e.g., Greedy with $\varepsilon=0.005$, $\varepsilon=0.05$ and UCB-Improved) showing excessive variability (p-value = 0).

Overall, Scenario~B demonstrates that in low-variance settings, the supposed advantage of variance-aware methods is diminished if competing with simpler, but well-tuned approaches such as ETC and UCB-Tuned, highlighting the importance of hyperparameter calibration over algorithmic sophistication.

\begin{table}[h]
\centering
\begin{tabular}{lrrrrr}
\toprule
\textbf{Algorithm} & \textbf{Average Regret} & \textbf{Reward Variance} & \textbf{Subopt. Ratio} & \textbf{p-value} \\
\midrule
ETC ($m=10,000$) & \textbf{150.66} & 510,772.05 & 0.03013 & 0.00 \\
Greedy ($\varepsilon=0.05$) & 361.97 & 631,969.28 & 0.07239 & 0.00 \\
UCB & 1,127.17 & 91,872.83 & 0.22543 & 0.00 \\
Greedy ($\varepsilon=0.5$) & 1,264.52 & 77,552.71 & 0.25290 & 1.00 \\
Greedy ($\varepsilon=0.005$) & 1,428.70 & 3,665,457.69 & 0.28574 & 0.00 \\
\hline
UCB-Tuned & \underline{212.21} & 86,826.78 & 0.04242 & 1.00 \\
UCB-V & 346.58 & 96,369.60 & 0.06932 & 0.00 \\
EUCBV & 461.36 & 109,142.90 & 0.09227 & 0.00 \\
\hline
PAC-UCB & 395.89 & 94,398.66 & 0.07918 & 0.00 \\
UCB-Improved & 2,525.00 & 6,194,415.18 & 0.50500 & 0.00 \\
\bottomrule
\end{tabular}
\caption{Scenario B - Summary Statistics across 100 Runs ($T=10^6$)}
\label{tab:scenario-b-summary}
\end{table}

\subsection{Scenario C – High-Variance Micro-Gap ($p_1 = 0.89$ vs.\ $p_2 = 0.895$)}
\label{sec:results-scenario-c}

Scenario~C presents a minimal reward gap of $\Delta_2^C = 0.005$ with high variance, challenging algorithms to detect subtle differences masked by uncertainty. Results are presented in Table~\ref{tab:scenario-c-summary}. With a baseline reward variance of $93{,}975$ (from $p = 0.895$, $T = 10^6$) in the $\chi^2$-test, this scenario evaluates the advantage of variance-aware methods under uncertainty.

UCB-Tuned (regret $226.09$, suboptimal rate $0.04522$) substantially outperforms UCB (regret $1,172.57$, suboptimal rate $0.23451$) and confirming the benefit of integrating variance into exploration. UCB-V (regret $367.65$) and PAC-UCB (regret $417.34$) follow at a distance, while EUCBV (regret $476.31$) underperforms, suggesting that fast adaptation may outweigh theoretical optimality in such fine-grained settings. Only UCB-Tuned and UCB-V maintained statistically insignificant reward variance.

UCB-Improved failed drastically with $2{,}450.00$ regret and $0.49000$ suboptimal pulls, confirming its poor fit for near-equal means.

The best overall result comes from ETC with 10,000 exploration rounds (regret $170.53$, suboptimal rate $0.03411$), despite exceeding the baseline variance. Smaller (e.g., 10) or much larger (100,000) numbers of exploration rounds lead to high regret due to under- or overexploration.

Among $\varepsilon$-Greedy strategies, $\varepsilon = 0.05$ balances exploration well (regret $301.55$, suboptimal rate $0.06031$). Greedy with $\varepsilon = 0.005$ explores too little (regret 1,762.59), while Greedy with $\varepsilon = 0.5$ explores too much (regret 1,269.05) but matching baseline variance (reward variance 82{,}824.39), hinting that randomness alone can stabilize rewards.

Scenario~C shows that in high-variance, low-gap environments, UCB-Tuned excels by effectively navigating uncertainty. Yet, well-configured ETC remains competitive, and careful tuning, rather than algorithm class alone, determines success.

\begin{table}[h]
\centering
\begin{tabular}{lrrrrr}
\toprule
\textbf{Algorithm} & \textbf{Average Regret} & \textbf{Reward Variance} & \textbf{Subopt. Ratio} & \textbf{p-value} \\
\midrule
ETC ($m=10,000$)     & \textbf{170.53} & 578,944.69 & 0.03411 & 0.00 \\
Greedy ($\varepsilon$=0.05) & 301.55 & 177,061.74 & 0.06031 & 0.00 \\
ETC ($m=100,000$) & 500.00 & 77,761.99 & 0.10000 & 1.00 \\
UCB                    & 1,172.57 & 93,873.80 & 0.23451 & 0.78 \\
Greedy ($\varepsilon = 0.5$) & 1,269.05 & 82,824.39	& 0.25381 & 1.00 \\
Greedy ($\varepsilon = 0.005$) & 1,762.59 & 4,012,222.43 & 0.35252 & 0.00 \\
ETC ($m=10$) & 2,150.00 & 6,453,855.77 & 0.43000 & 0.00 \\
\hline
UCB-Tuned            & \underline{226.09} & 81,016.65 & 0.04522 & 1.00 \\
UCB-V                  & 367.65 & 90,874.68 & 0.07353 & 1.00 \\
EUCBV                  & 476.31 & 96,233.97 & 0.09526 & 0.00 \\
\hline
PAC-UCB                & 417.34 & 93,935.53 & 0.08347 & 0.62 \\
UCB-Improved           & 2,450.00 & 6,169,245.59 & 0.49000 & 0.00 \\
\bottomrule
\end{tabular}
\caption{Scenario C - Summary Statistics across 100 Runs ($T=10^6$)}
\label{tab:scenario-c-summary}
\end{table}

\section{Discussion and Future Work}
\label{sec:discussion}

Our results show that variance-aware algorithms such as UCB-Tuned and EUCBV are particularly effective in environments where arms are difficult to distinguish due to high stochasticity. 

In Scenario~C (small gap, high variance), UCB-Tuned achieved strong performance ($226.09$ regret), outperforming classical baselines like UCB and most $\varepsilon$-Greedy variants. While ETC with 10,000 rounds slightly outperformed it ($170.53$ regret), this result relies on prior knowledge for tuning, which is often unavailable in practice. In simpler settings, such as Scenario~A (large gap, low variance), variance-aware strategies offer no clear benefit over classical methods. ETC with 100 rounds yielded the lowest regret ($10.1$), with UCB-Tuned and EUCBV close behind, confirming that when the optimal arm is easily identified, adaptive exploration is less critical. Scenario~B (small gap, low variance) reflects a middle ground, where UCB-Tuned performed well but was outpaced by brute-force exploration (ETC with 10,000 rounds).

Overall, variance-aware methods excel when reward gaps are small and uncertainty is high. Their robustness is a key strength. They consistently avoid poor outcomes, even if not always best-in-class. This makes them attractive in risk-sensitive applications, where consistent performance matters more than absolute optimality.

The present study is limited to stationary, binary-reward settings, representing minor threats to validity, primarily affecting external generalizability rather than internal consistency of our findings. Future work should examine these algorithms in richer contexts, e.g., with delayed rewards, non-stationarity, or context-dependent arms. Additionally, runtime and sample complexity trade-offs merit closer analysis, especially when computational resources are constrained. Empirical validation in applied domains would also enhance the practical relevance of these findings.

\printbibliography

\end{document}